\documentclass[10pt,twocolumn,letterpaper]{article}

\usepackage[pagenumbers]{cvpr} 

\usepackage{graphicx}
\usepackage{amsmath}
\usepackage{amssymb}
\usepackage{booktabs}
\usepackage{color, colortbl}
\usepackage[dvipsnames]{xcolor}
\usepackage[accsupp]{axessibility}

\definecolor{citecolor}{HTML}{0071bc}
\usepackage[pagebackref,breaklinks=true,letterpaper=true,colorlinks,citecolor=citecolor,bookmarks=false]{hyperref}

\usepackage[capitalize]{cleveref}
\crefname{section}{Sec.}{Secs.}
\Crefname{section}{Section}{Sections}
\Crefname{table}{Table}{Tables}
\crefname{table}{Tab.}{Tabs.}

\usepackage[ruled,lined,linesnumbered,algo2e]{algorithm2e}

\renewcommand{\paragraph}[1]{\noindent\textbf{#1}}
\renewcommand{\vec}{\boldsymbol}
\usepackage{gensymb}
\usepackage{booktabs}

\let\oldsubsection\subsection
\renewcommand{\subsection}[1]{\vspace{-1mm}\oldsubsection{#1}\vspace{-1mm}}

\newcommand{\reffig}[1]{Figure~\ref{fig:#1}}

\newcommand{\refsec}[1]{Section~\ref{sec:#1}}

\newcommand{\reftbl}[1]{Table~\ref{tbl:#1}}

\newcommand{\refeq}[1]{Equation~\eqref{eq:#1}}

\newcommand{\lblfig}[1]{\label{fig:#1}}
\newcommand{\lblsec}[1]{\label{sec:#1}}
\newcommand{\lbleq}[1]{\label{eq:#1}}

\newcommand{\lbltbl}[1]{\label{tbl:#1}}

\newcommand{\rowNumber}[1]{\textcolor{Cerulean}{#1}}

\def\cvprPaperID{***}
\def\confName{CVPR}
\def\confYear{2022}

\begin{document}

\title{Global Tracking Transformers}

\author{
  Xingyi Zhou$^{1}$ \quad Tianwei Yin$^{1}$ \quad Vladlen Koltun$^2$ \quad Philipp Kr{\"a}henb{\"u}hl$^1$\\
{$^1$The University of Texas at Austin \quad \quad $^2$Apple}
}

\maketitle

\begin{abstract}
We present a novel transformer-based architecture for global multi-object tracking. Our network takes a short sequence of frames as input and produces global trajectories for all objects. The core component is a global tracking transformer that operates on objects from all frames in the sequence. The transformer encodes object features from all frames, and uses trajectory queries to group them into trajectories. The trajectory queries are object features from a single frame and naturally produce unique trajectories. Our global tracking transformer does not require intermediate pairwise grouping or combinatorial association, and can be jointly trained with an object detector. It achieves competitive performance on the popular MOT17 benchmark, with $75.3$ MOTA and $59.1$ HOTA. More importantly, our framework seamlessly integrates into state-of-the-art large-vocabulary detectors to track any objects. Experiments on the challenging TAO dataset show that our framework consistently improves upon baselines that are based on pairwise association, outperforming published works by a significant $7.7$ tracking mAP. Code is available at \url{https://github.com/xingyizhou/GTR}.
\end{abstract}

\section{Introduction}

Multi-object tracking aims to find and follow all objects in a video stream.
It is a basic building block in application areas such as mobile robotics, where an autonomous system must traverse dynamic environments populated by other mobile agents.
In recent years, \emph{tracking-by-detection} has emerged as the dominant tracking paradigm, powered by advances in deep learning and object detection~\cite{ren2015faster,he2017mask}.
Tracking-by-detection reduces tracking to two steps: detection and association.
First, an object detector independently finds potential objects in each frame of the video stream.
Second, an association step links detections through time.
Local trackers~\cite{Bewley2016_sort,Wojke2018deep,bergmann2019tracking,zhang2020fair,wang2019towards,xu2019spatial} primarily consider pairwise associations in a greedy way (\reffig{teaser:local}).
They maintain a status of each trajectory based on location~\cite{zhou2020tracking,Bewley2016_sort} and/or identity features~\cite{zhang2020fair,Wojke2018deep},
and associate current-frame detections with each trajectory based on its last visible status.
This pairwise association is efficient, but lacks an explicit model of trajectories as a whole, and sometimes struggles with heavy occlusion or strong appearance change.
Global trackers~\cite{zhang2008global,tang2017multiple,braso2020learning,yu2007multiple,berclaz2011multiple} run offline graph-based combinatorial optimization over pairwise associations.
They can resolve inconsistently grouped detections and are more robust, but can be slow and are usually detached from the detector.
\begin{figure}[!t]
\centering
    \begin{subfigure}{\linewidth}
    \includegraphics[page=1, width=0.99\linewidth]{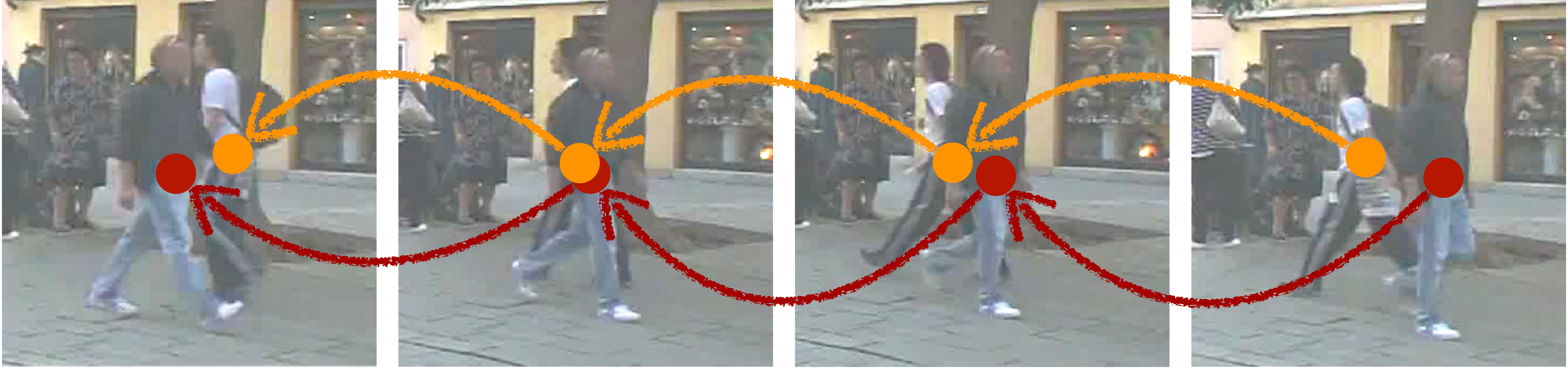}
    \caption{Local trackers}
    \lblfig{teaser:local}
    \end{subfigure}
    \begin{subfigure}{\linewidth}
    \includegraphics[page=2, width=0.99\linewidth]{figs/teaser3.pdf}
    \caption{Our global tracker}
    \lblfig{teaser:ours}
    \vspace{-3mm}
    \end{subfigure}
   \caption{
   \textbf{Local trackers (top) vs. our global tracker (bottom).} 
   Local trackers associate objects frame-by-frame, optionally with a external track status memory (not show in the figure). Our global tracker take a short video clip as input, and associates objects across all frames using global object queries.
   }
\lblfig{teaser}
\vspace{-5mm}
\end{figure}

\begin{figure*}[!t]
\centering
   \includegraphics[page=2,width=0.9\linewidth]{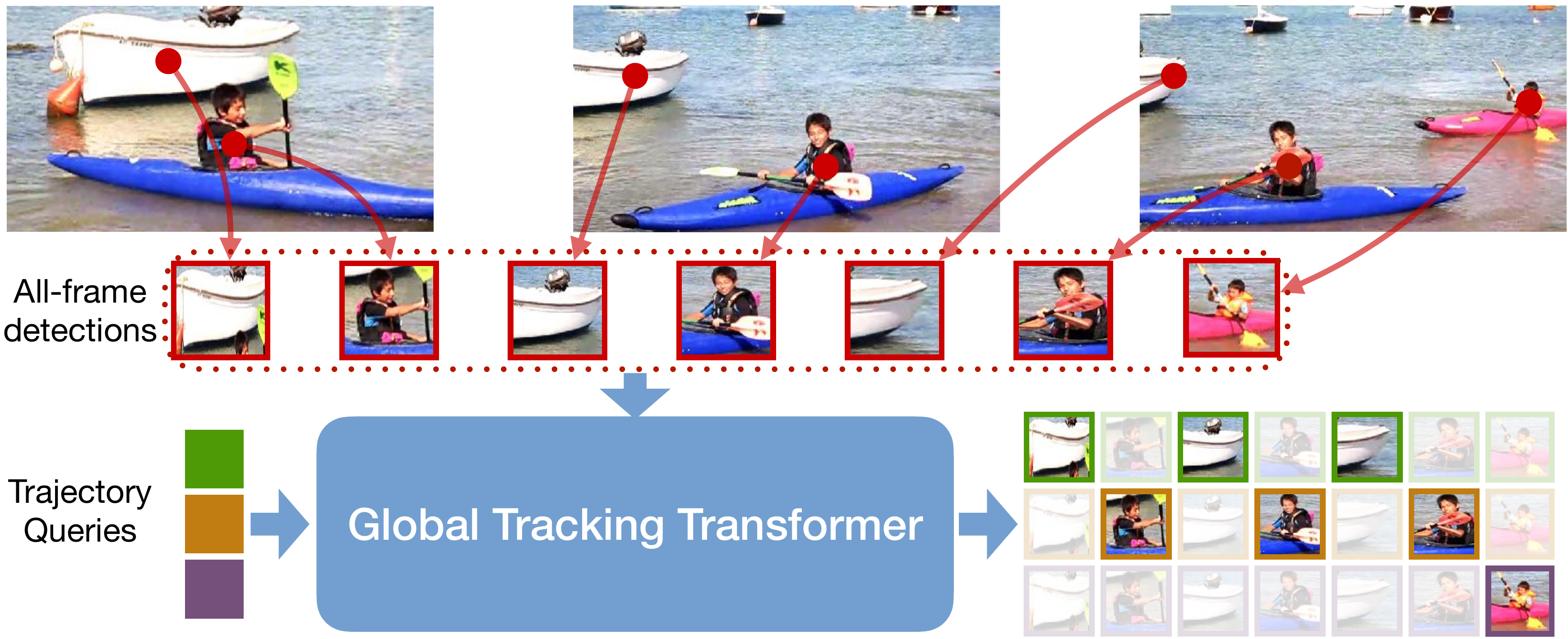}
   \caption{
   \textbf{Overview of our joint detection and tracking framework.} 
   An object detector first independently detects objects in all frames. Object features are concatenated and fed into the encoder of our global Tracking transformer (GTR). 
   The GTR additionally takes trajectory queries as decoder input, and produces association scores between each query and object.
   The association matrix links objects for each query.
   During testing, the trajectory queries are object features in the last frame.
   The structure of the transformer is shown in \reffig{structure}.
   }
\lblfig{framework}
\end{figure*}

In this work, we show how to represent global tracking (\reffig{teaser:ours}) as a few layers in a deep network (\reffig{framework}).
Our network directly outputs trajectories and thus sidesteps both pairwise association and graph-based optimization.
We show that detectors~\cite{ren2015faster,he2017mask,zhou2019objects} can be augmented by transformer layers to turn into joint detectors and trackers.
Our Global TRacking transformer (GTR) encodes detections from multiple consecutive frames, and uses \emph{trajectory queries} to group them into trajectories.
The queries are detection features from a single frame 
(e.g., the current frame in an online tracker) 
after non-maximum suppression, and are transformed by the GTR into trajectories.
Each trajectory query produces a single global trajectory by assigning to it a detection from each frame using a softmax distribution.
The outputs of our model are thus detections and their associations through time.
During training, we explicitly supervise the output of our global tracking transformer using ground-truth trajectories and their image-level bounding boxes.
During inference, we run GTR in a sliding window manner with a moderate temporal size of $32$ frames, and link trajectories between windows online.
The model is end-to-end differentiable within the temporal window.

Our framework is motivated by the recent success of transformer models~\cite{vaswani2017attention} in computer vision in general~\cite{ZhaoJK20,dosovitskiy2020image,touvron2020deit,liu2021swin} and in object detection in particular~\cite{carion2020end,wang2020end}.
The cross-attention structure between queries and encoder features mines similarities between objects and naturally fits the association objective in multi-object tracking.
We perform cross-attention between trajectory queries and object features within a temporal windows, and explicitly supervise it to produce a query-to-detections assignment.
Each assignment directly corresponds to a global trajectory.
Unlike transformer-based detectors~\cite{carion2020end, wang2020end,meinhardt2021trackformer,transtrack} that learn queries as fixed parameters, our queries come from existing detection features and adapt with the image content.
Furthermore, our transformer operates on detected objects rather than raw pixels~\cite{carion2020end}.
This enables us to take full advantage of well-developed object detectors~\cite{zhou2021probablistic,he2017mask}.

Our framework is end-to-end trainable, and easily integrates with state-of-the-art object detectors.
On the challenging large-scale TAO dataset, our model reaches $20.1$ tracking mAP on the test set, significantly outperforming published work, which achieved $12.4$ tracking mAP~\cite{qdtrack}.
On the MOT17~\cite{MOT16} benchmark, our entry achieves competitive $75.3$ MOTA and $59.1$ HOTA, outperforming most concurrent transformer-based trackers~\cite{meinhardt2021trackformer,xu2021transcenter,zeng2021motr}, and on-par with state-of-the-art association-based trackers.

\section{Related work}

\paragraph{Local multi-object tracking.}
Many popular trackers operate locally and greedily~\cite{Bewley2016_sort,Wojke2018deep,bergmann2019tracking,zhou2020tracking,xu2021transcenter,wang2019towards,zhang2020fair,Tokmakov_2021_ICCV}.
They maintain a set of confirmed tracks, and link newly detected objects to tracks based on pairwise object-track distance metrics.
SORT~\cite{Bewley2016_sort} and DeepSORT~\cite{Wojke2018deep} model tracks with Kalman filters, and update the underlying locations~\cite{Bewley2016_sort} or deep features~\cite{Wojke2018deep} in every step.
Tracktor~\cite{bergmann2019tracking} feeds tracks to a detector as proposals, and directly propagates the tracking ID.
CenterTrack~\cite{zhou2020tracking} conditions detection on existing tracks, and associates objects using their predicted locations.
TransCenter~\cite{xu2021transcenter} builds upon CenterTrack by incorporating deformable DETR~\cite{zhu2020deformable}.
JDE~\cite{wang2019towards} and FairMOT~\cite{zhang2020fair} train the detector together with an instance-classification branch, and associate via pairwise ReID features similar to SORT~\cite{Bewley2016_sort}.
STRN~\cite{xu2019spatial} learns a dedicated association feature considering the spatial and temporal cues, but again performs pairwise association.
In contrast, we do not rely on pairwise association, but instead 
associates to all objects across the full temporal window via a transformer.

\paragraph{Global tracking.}
Traditional trackers first detect objects offline, and consider object association across all frames as a combinatorial optimization problem~\cite{peng2020tpm,zhang2008global,braso2020learning,dai2021learning,tang2017multiple}.
Zhang et al.~\cite{zhang2008global} formulate tracking as a min-cost max-flow problem over a graph, where nodes are detections and edges are valid associations.
MPN~\cite{braso2020learning} simplifies the graph construction and proposes a neural solver that performs the graph optimization.
LPC~\cite{dai2021learning} additionally considers a classification module on the graph.
Lif\_T~\cite{tang2017multiple} incorporates person ReID and pose features in the graph optimization.
These methods are still based on pairwise associations and use the combinatorial optimization to select globally consistent 
assignments.
Our method directly outputs consistent long-term trajectories without combinatorial optimization.
This is done by a single forward pass in a relatively shallow network.

\paragraph{Transformers in tracking.}
Trackformer~\cite{meinhardt2021trackformer} augments DETR~\cite{carion2020end} with additional object queries from existing tracks,
and propagates track IDs as in Tracktor~\cite{bergmann2019tracking}.
TransTrack~\cite{transtrack} uses features from historical tracks as queries, but associates objects based on updated bounding box locations.
MOTR~\cite{zeng2021motr} follows the DETR~\cite{carion2020end} structure and iteratively propagates and updates track queries to associate object identities.
MO3TR~\cite{zhu2021looking} additionally uses a temporal attention module to update the status of each track over a temporal window, and feeds updated track features as queries in DETR.
The common idea behind these works is to use the object query mechanism in DETR~\cite{carion2020end} to extend existing tracks \emph{frame-by-frame}.
We use transformers in a different way.
Our transformer uses queries to generate entire trajectories at once.
Our queries do not generate new boxes, but group already-detected boxes into trajectories.

\textbf{Video object detection.}
Applying attention blocks on object features over a video is a successful idea in video object detection~\cite{russakovsky2015imagenet}.
SELSA~\cite{wu2019sequence} feeds region proposals of randomly sampled frames to a self-attention block to provide global context.
MEGA~\cite{chen2020memory} builds a hierarchical local and global attention mechanism with a large temporal receptive field.
ContextRCNN~\cite{beery2020context} uses an offline long-term feature bank~\cite{wu2019long} to integrate long-range temporal features.
These methods support our idea of using transformers to analyze object relations.
The key difference is they do not use object identity information, but implicitly use object correlations to improve detection.
We explicitly learn object association in a supervised way for tracking.

\section{Preliminaries}

We start by formally defining object detection, tracking, and tracking by detection.

\paragraph{Object detection.}
Let $I$ be an image.
The goal of object detection is to identify and localize all objects.
An object detector~\cite{zhou2019objects,tian2019fcos,ren2015faster,carion2020end} takes the image $I$ as input and produces a set of objects $\{p_i\}$ with locations $\{b_i\}, b_i \in \mathbb{R}^4$ as its output.
For multi-class object detection, a second stage~\cite{ren2015faster,he2017mask} takes the object features and 
produce a classification score $s_i \in \mathbb{R}^C$ from a set of predefined classes $C$ and a refined location $\tilde b_i$.
For single-class detection (e.g., pedestrian detection~\cite{MOT16}), the second stage can me omitted. 

\paragraph{Tracking.} Let $I^1, I^2, \ldots, I^T$ be a series of images.
The goal of a tracker is to find trajectories $\vec \tau_1, \vec \tau_2, \ldots, \vec \tau_K$ of all objects over time.
Each trajectory $\vec \tau_k=[\tau_k^1, \ldots, \tau_k^T]$ describes a tube of object locations $\tau_k^t \in \mathbb{R}^4 \cup\{\emptyset\}$ through time $t$.
$\tau_k^t = \emptyset$ indicates that the object $k$ cannot be located in frame $t$.
The tracker may optionally predict the object class score $s_k$~\cite{dave2020tao} for each trajectory, usually as the average class of its per-frame slices.

\paragraph{Tracking by detection} decomposes the tracking problem into per-frame detection and inter-frame object association.
Object detection first finds $N_t$ candidate objects $b^t_1, b^t_2, \ldots$ as bounding boxes $b^t_i \in \mathbb{R}^4$ in each frame $I^t$.
Association then links existing tracks $\vec \tau_k$ to current detected objects using an object indicator $\alpha_k^t \in \{\emptyset, 1, 2, \ldots, N_t\}$ at each frame $t$ :
$$
 \tau_k^t = \begin{cases} \emptyset & \text{if $\alpha_k^t = \emptyset$}\\b^t_{\alpha_k^t} &\text{otherwise} \end{cases}
$$

Most prior works define the association greedily through pairwise matches between objects in adjacent or nearby frames~\cite{Bewley2016_sort,bergmann2019tracking,zhou2020tracking,zhang2020fair}, or rely on offline combinatorial optimization for global association~\cite{zhang2008global,braso2020learning,frossard2018end}.
In this work, we show how to perform joint detection and global association within a single forward pass through a network.
The network learns global tracking within a video clip of $32$ frames in an end-to-end fashion.
We leverage a probabilistic formulation of the association problem and show how to instantiate tracking in a transformer architecture in \refsec{main}.

\section{Global tracking transformers}
\lblsec{main}
Out global tracking transformer (GTR) associates objects in a probabilistic and differentiable manner.
It links objects $p_i^t$ in each frame $I^t$ to a set of trajectory queries $q_k$.
Each trajectory query $q_k$ produces an object association score vector $g \in \mathbb{R}^N$ over objects from all frames.
This association score vector then yields a per-frame object-level association $\alpha^t_k \in \{\emptyset, 1,\ldots,N_t\}$, where $\alpha^t_k=\emptyset$ indicates no association and $N_t$ is the number of detected objects in frame $I^t$.
The combination of associations then produces a trajectory $\vec \tau_k$.
\reffig{framework} provides an overview.
The association step is differentiable and can be jointly trained with the underlying object detector.

\subsection{Tracking transformers}
Let $p_1^t, \ldots, p_{N_t}^t$ be a set of high-confidence objects for image $I^t$.
Let $B^t=\{b_1^t, \ldots, b_{N_t}^t\}$ be their corresponding bounding boxes.
Let $f_i^t \in \mathbb{R}^{D}$ be the $D$-dimentional features
extracted from boxes $b_i^t$.
For convenience let $F^t = \{f_1^t, \ldots, f_{N_t}^t\}$ be the set of all detection features of image $I^t$, and $F = F^1 \cup \ldots \cup F^T$ be the set of all features through time.
The collection of all object features $F \in \mathbb{R}^{N \times D}$ is the input to our tracking transformer, where $N = \sum_t^T N_t$ is the total number of detections in all frames.
The tracking transformer takes features $F$ and a trajectory query $q_k \in \mathbb{R}^D$, and produces a trajectory-specific association score $g(q_k, F) \in \mathbb{R}^{N}$.

Formally, let $g_i^t(q_k, F) \in \mathbb{R}$ be the score of the $i$-th object in the $t$-th frame.
A special output token $g_\emptyset^t(q_k, F)=0$ indicates no association at time $t$.
The tracking transformer then predicts a distribution of associations over all objects $i$ in frame $I^t$  for each trajectory $k$.
We model this as an independent softmax activation for each time-step $t$:
\begin{equation}
    P_A(\alpha^t = i | q_k, F) = \frac{\exp\left(g_{i}^t(q_k, F)\right)}{\sum_{j\in\{\emptyset, 1, \ldots N_t\}}\exp\left(g_j^t(q_k, F)\right)}
    \lbleq{asso_cost}
\end{equation}
Since a detector produces a single bounding box $b_i^t$ for each object $p_i^t$, there is a one-to-one mapping between the association distribution $P_A$ and a distribution $P_t$ over bounding boxes for trajectory $k$ at time $t$:
$
P_t(b|q_k, F) = \sum_{i=1}^{N_t} 1_{[b = b_i^t]} P_A(\alpha^t = i | q_k, F),
$
where the indicator $1_{[\cdot]}$ assigns an output bounding box to each associated query.
In practice, a detector's non-maximum suppression (NMS) ensures that there is also a unique mapping from $P_t$ back to $P_A$.
The distribution over bounding boxes in turn leads to a distribution over entire trajectories 
${P_T(\vec \tau|q_k, F) = \prod_{t=1}^T P_t(\tau^t|q_k, F)}$.

During training, we maximize the $\log$-likelihood of the ground-truth trajectories.
During inference, we use the likelihood to produce long-term tracks in an 
online manner.

\subsection{Training}
Given a set of ground-truth trajectories $\vec{\hat \tau_1}, \ldots, \vec{\hat \tau_K}$, our goal is to learn a tracking transformer that estimates $P_A$, and implicitly the trajectory distribution $P_T$.
We jointly train the tracking transformer with detection by treating the transformer as 
an RoI head like two-stage detectors~\cite{ren2015faster}.
At each training iteration, we first obtain high-confidence objects $b_1^t, \ldots, b_{N_t}^t$ and their corresponding features $F_t$ after non-maximum suppression.
We then maximize $\log P_T(\vec \tau |q_k, F)$ for each ground-truth trajectory $\vec \tau$.
This is equivalent to maximizing $\log P_A(\alpha^t|q_k, F)$ after assigning $\vec \tau$ to a set of objects.
We follow object detection and use a simple intersection-over-union (IoU) assignment rule:
\begin{equation}
  \hat \alpha_k^t\!=\!\begin{cases} 
  \emptyset, \quad \quad \text{if $\tau_k^t\!=\!\emptyset $ or  $\max_i IoU(b_i^t, \tau_k^t)\!<\!0.5$}\\
  { \text{argmax}_i IoU(b_i^t, \tau_k^t)}, \text{otherwise} 
  \end{cases}
  \lbleq{assign}
\end{equation}
We use this assignment to both train the bounding box regression of the underlying two-stage detector, and our assignment likelihood $P_A$.
However, this assignment likelihood further depends on the trajectory query $q_k$, which we define next.

\paragraph{Trajectory queries}
\lblsec{queries}
are key to our formulation. Each query $q_k$ generates a trajectory.
In prior work~\cite{carion2020end}, object queries were learned as network parameters and fixed during inference.
This makes queries image-agnostic and requires a near-exhaustive enumeration of them~\cite{zhu2020deformable,sun2020rethinking,liu2021group}.
For objects this is feasible~\cite{carion2020end}, as anchors~\cite{lin2017focal} or proposals~\cite{sun2020sparse} showed.
Trajectories, however, live in the exponentially larger space of potential moving objects than simple boxes, and hence require many more queries to cover that space.
Furthermore, tracking datasets feature many fewer annotated instances, and learned trajectories easily overfit and remember the dataset.

We instead directly use object features $f_i^t$ as the object queries.
Specifically, let $\hat \alpha_k$ be the matched objects for a ground-truth trajectory $\vec \tau_k$ according to \refeq{assign}.
Any feature $\{f_{\hat\alpha_k^1}^1, f_{\hat\alpha_k^2}^2, \ldots\}$ can serve as the trajectory query for trajectory $\vec \tau_k$.
In practice, we use all object features $F$ in the all the $T$ frames as queries and train the transformer for a sequence length of $T$.
Any unmatched features $f_i^t$ are used as background queries and supervised to produce $\emptyset$ for all frames.
We allow multiple queries to produce the same trajectory, and do not require a one-to-one match~\cite{carion2020end}.
During inference, we only use object features from one single frame as the queries to avoid duplicate outputs.
All object features within a frame are different (after standard detection NMS) and hence produce different trajectories.

\paragraph{Training objective.}
The overall training objective combines the assignment in \refeq{assign} and trajectory queries to maximize the log-likelihood of each trajectory under its assigned queries.
For each trajectory $\tau_k$ we optimize the log-likelihood of its assignments $\hat \alpha_k$:
\begin{equation}
\ell_{asso}(F, \vec{\hat \tau}_k)\!=\!-\!\!\!\!\!\!\sum_{s \in \{1 \ldots T | \hat\alpha_k^s \ne \emptyset\}}\sum_{t=1}^T\log P_A(\hat \alpha_k^t | F^s_{\hat \alpha_k^s}, F)
\end{equation}
For any unassociated features, we produce empty trajectories:
\begin{equation}
\ell_{bg}(F) = -\sum_{s=1}^T \sum_{j: \nexists\hat \alpha_k^s = j}\sum_{t=1}^T\log P_A(\alpha^t=\emptyset | F^s_j, F)
\end{equation}
The final loss simply combines the two terms:
\begin{equation}
  L_{asso}(F, \{\vec{\hat \tau}_1,\ldots,\vec{\hat \tau}_K\})\!=\!\ell_{bg}(F)\!+\!\sum_{\vec{\hat \tau}_k} \ell_{asso}(F,\vec{\hat \tau}_k)
\end{equation}
We train $L_{asso}$ jointly with standard detection losses~\cite{zhou2019objects}, including classification and bounding-box regression losses, and 
optionally second stage classification and regression losses for multi-class tracking~\cite{dave2020tao}, 

\begin{figure*}[!t]
\centering
   \includegraphics[width=0.95\linewidth]{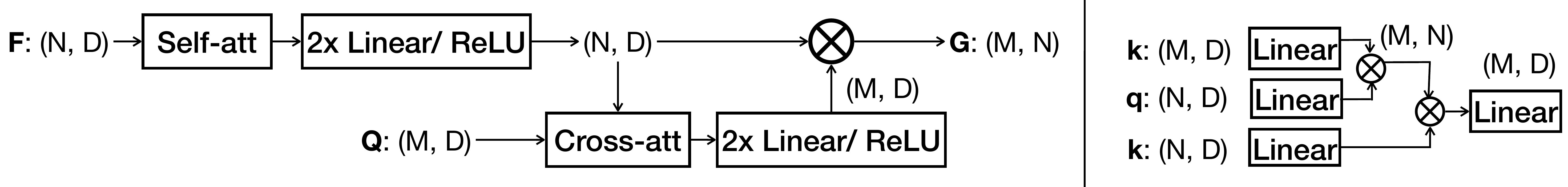}
   \vspace{-3mm}
   \caption{Left: detailed network architecture of GTR. Right: detailed structure of both self-att and cross-att blocks. We omit multi-head~\cite{vaswani2017attention} in the figure for simplicity. For self-attention, $q = k = F$. For cross attention, $q = Q$, $k = F$. We list data dimensionalities in parentheses. $\bigotimes$ indicates matrix multiplication (transpose when needed).}
\lblfig{structure}
\vspace{-5mm}
\end{figure*}

\subsection{Online Inference}
\lblsec{inference}

During inference, we process the video stream online in a sliding-window manner with window size $T=32$ and stride $1$.
For each individual frame $t$, we feed the image to the network before the tracking transformer and obtain $N_t$ bounding boxes $B^t$ and object features $F^t$.
We keep a temporal history buffer of $T$ frames, i.e., $B = \{B^{t - T + 1}, \cdots, B^{t}\}$ and $F = \{F^{t - T + 1}, \cdots, F^{t}\}$, and run the tracking transformer for each sliding window.
We use object features from the current frame $t$ as trajectory queries $q_k = F^t_k$ to produce $N_t$ trajectories.
For the first frame, we initialize all detections as trajectories.
For any subsequent frame, we link current predicted trajectories to existing tracks using the average assignment likelihood $P_A$ as a distance metric.
Since the current trajectories share up to $T-1$ boxes and features with past trajectories, the overlap can be quite large.
We use a Hungarian algorithm to ensure that the mapping from current long-term trajectories to existing tracks is unique.
If the average association score with any prior trajectory is lower than a threshold $\theta$, we start a new track.
Otherwise we append the underlying current detection (query) that generates the trajectory to the matched existing track.

\subsection{Network architecture}
The global tracking transformer takes a stack of object features $F\in \mathbb{R}^{N\times D}$ as the encoder input, a matrix of queries $Q \in \mathbb{R}^{M\times D}$ as the decoder input, and produces an association matrix $G \in \mathbb{R}^{M \times N}$ between queries and objects.
The detailed structure of the tracking transformer is shown in \reffig{structure} (left).
It follows DETR~\cite{carion2020end} but only uses a one-layer encoder and a one-layer decoder.
Empirically, we observe that self-attention for queries and Layer Normalization~\cite{ba2016layer} were not required.
See \refsec{ablation} for an ablation.
The resulting network structure is lightweight, with 10 linear layers in total.
It runs in a fraction of the runtime of the backbone detector, even for hundreds of queries.

\subsection{Connection to embedding learning and ReID}
\lblsec{connection}
Consider a variation of GTR with just a dot-product association score $g_i^t(q_k, F) = q_k \cdot F_i^t$.
Further consider learning all trajectory queries $Q = \left\{q_1, \ldots, q_k\right\}$ as free parameters, one per training trajectory $\tau_k$.
In this variation, the softmax assignment in \refeq{asso_cost} reduces to a classification problem.
For each object feature, we classify it as a specific training instance or as background.
This is exactly the objective of classification-based embedding learning in person-ReID~\cite{Luo_2019_CVPR_Workshops}, as used in ReID-based trackers~\cite{zhang2020fair,wang2019towards}.

The two key differences between embedding learning and GTR are: first, our transformer does not assume any factorization of $g_i^t$ and allows the model to reason about all boxes at once when computing associations.
A dot-product-based ReID network on the other hand assumes that all boxes independently produce a compatible embedding.
See \refsec{ablation} for an ablation of this transformer structure.
Second, our trajectory queries are not learned.
This allows our transformer to produce long-term associations in a single forward pass, while ReID-based trackers rely on a separate cosine-distance-based grouping step~\cite{zhang2020fair,wang2019towards}.

\section{Experiments}
We evaluate our method on two tracking benchmarks: TAO~\cite{dave2020tao} and MOT17~\cite{MOT16}.

\textbf{TAO}~\cite{dave2020tao} tracks a wide variety of objects.
The images are adopted from 6 existing video datasets, including indoor, outdoor, and driving scenes. 
The dataset requires tracking objects with a large vocabulary of 488 classes in a long-tail setting.
It contains 0.5k, 1k, and 1.5k videos for training, validation, and testing, respectively.
Each video contains $\sim40$ annotated frames at 1 annotated frame per second.
There is significant motion between adjacent annotated frames.
The training annotations are incomplete.
We thus do not use the training set and solely train on LVIS~\cite{gupta2019lvis} and use the TAO validation and test set for evaluation.

\textbf{MOT}~\cite{MOT16} tracks pedestrians in crowd scenes. 
It contains 7 training sequences and 7 test sequences. The sequences contain 500 to 1500 frames, recorded and annotated at 25-30 FPS. 
We follow CenterTrack~\cite{zhou2020tracking} and split each training sequence in half.
We use the first half for training and the second half for validation.
We perform ablation studies mainly on this validation set, and compare to other approaches on the official hidden test set.
We evaluate under the private detection protocol. 

\subsection{Evaluation metrics}
We evaluate under the official metrics for each dataset.
TAO~\cite{dave2020tao} uses tracking mAP@0.5 as the official metric, which is based on standard object detection mAP~\cite{lin2014microsoft} but changes the 2D bounding box IoU to 3D temporal-spatial IoU between the predicted trajectory and the ground-truth trajectory. The overall tracking mAP is averaged across all classes.
MOT~\cite{MOT16} uses Multi-Object Tracking Accuracy (MOTA) as the official metric.
${\textrm{MOTA} = 1 - \frac{\sum_t(FP_t + FN_t + IDSW_t)}{\sum_t GT_t}}$,
where $GT_t$ is the number of ground truth objects in frame $t$, and $FP_t$, $FN_t$, and $IDSW_t$ measure the errors of false positives, false negatives, and ID switches, respectively.

As suggested by the MOT benchmark, we additionally report HOTA, a new tracking metric~\cite{luiten2021hota}.
HOTA is defined as the geometric mean of detection accuracy (DetA) and association accuracy (AssA).
Both DetA and AssA have the form $\frac{|TP|}{|TP| + |FN| + |FP|}$, with their respective true/false criteria.
In our experiments, we mainly use AssA to access tracking performance.

\subsection{Training and inference details}
\lblsec{details}

\begin{table*}[!t]
\centering
\begin{tabular}{@{}l@{\ \ \ }l@{\ \ \ }c@{\ \ \ }c@{\ \ \ }c@{\ \ \ }c@{\ \ \ }c@{\ \ \ }c@{\ \ }c@{\ \ \ }c@{\ \ \ }c@{}}
\toprule
\rowNumber{\#} & & \multicolumn{4}{c}{TAO} & \multicolumn{5}{c}{MOT17} \\ 
 & & Track mAP & HOTA & DetA & AssA  & MOTA & IDF1 & HOTA & DetA & AssA \\
\cmidrule(r){1-2}
\cmidrule(r){3-6}
\cmidrule(r){7-11}
\rowNumber{1} & IoU ~\cite{Bewley2016_sort} & 8.8 & 32.7 & 30.5 & 35.4 & 68.9 & 65.0 & 57.4 & 59.2 & 56.1\\
\rowNumber{2} & ReID & 10.9 & 35.0 & 31.4 & 39.5 & 70.9 & 74.0 & 61.7 & 60.0 & 63.7 \\
\rowNumber{3} & IoU+ReID ~\cite{zhang2020fair} & 11.0 & 34.9 & 31.2 & 39.5 & 71.1 & 74.2 & 62.1 & 60.2 & 64.4 \\
\rowNumber{4} & IoU+ReID (retrained) & 6.7 & 23.4 & 18.8 & 29.5 & 69.9 & 73.0 & 60.9 & 59.4 & 62.5  \\
\cmidrule(r){1-2}
\cmidrule(r){3-6}
\cmidrule(r){7-11}
\rowNumber{5} & GTR (T=2) & 13.6 & 42.0 & 35.8 & 49.8 & 71.3 & 65.1 & 57.8 & 60.6 & 55.8 \\
\rowNumber{6} & GTR (T=4) & 17.7 & 44.0 & 36.4 & 53.6 & \bf 71.6 & 69.6 & 59.9 & \bf 60.8 & 59.6 \\ 
\rowNumber{7} & GTR (T=8) & 19.5 & 45.6 & 36.8 & 56.8 & 71.3 & 72.2 & 61.1 & 60.7 & 62.0 \\
\rowNumber{8} & GTR (T=16) & {\bf 22.5} & {\bf 45.8} & {\bf 36.8} & {\bf 57.4} & 71.4 & 75.1 & 62.5 & 60.6 & 65.0 \\ 
\rowNumber{9} & GTR (T=32) & 22.1 & 44.9 & 35.9 & 56.7 & 71.3 & {\bf 75.9} & {\bf 63.0} & 60.4 & {\bf 66.2} \\
\bottomrule
\end{tabular}
\vspace{-3mm}
\caption{
\textbf{Effectiveness of global tracking.}
We compare greedy trackers~\cite{Bewley2016_sort,zhang2020fair} (top block) with our global tracker(GTR) under different temporal windows on the TAO and MOT17 validation sets. 
We show the official metrics (Track mAP for TAO and MOTA/ IDF1 for MOT17) as well as HOTA metrics. All metrics are higher better.
All rows except row \rowNumber{4} are the same model trained with our losses (evaluated with different tracking algorithms). 
Row \rowNumber{4} is a different model retrained with the original instance-classification loss~\cite{zhang2020fair}.
Our global tracker benefits from longer temporal windows, and outperforms local trackers.
}
\lbltbl{global}
\vspace{-5mm}
\end{table*}

\par \noindent \textbf{TAO training.}
Our implementation is based on detectron2~\cite{wu2019detectron2}.
For TAO~\cite{dave2020tao} experiments, we use Res2Net~\cite{gao2019res2net} with deformable convolution~\cite{dai2017deformable} as the backbone.
We adopt CenterNet2~\cite{zhou2021probablistic} as the detector, which uses CenterNet~\cite{zhou2019objects} as proposal network and 
cascaded RoI heads~\cite{cai2018cascade} for classification.
Following the guideline of TAO dataset~\cite{dave2020tao}, we train the object detector on the combination of LVISv1~\cite{gupta2019lvis} and COCO~\cite{lin2014microsoft}.
We additionally incorporate a federated loss~\cite{zhou2021probablistic} to improve long-tail detection.
We first train a single-frame detector.
The training uses SGD with learning rate $0.04$ and batch size 32 for 180K iterations (the $4\!\times$ schedule~\cite{wu2019detectron2}).
We use training resolution $896\!\times\!896$ following the scale-and-crop augmentation of EfficientDet~\cite{tan2020efficientdet}.
The detector yields $37.1$ mAP on the LVISv1 validation set and $27.3$ mAP on the TAO validation set.

TAO only provides a small training set for tuning tracking hyperparameters, but not for training the tracker.
We empirically observed that training on the TAO training set hurts detection performance, and overall does not yield good tracking accuracy.
We find that training only on static image datasets~\cite{gupta2019lvis} with data augmentation is sufficient for tracking.
Our training strategy follows CenterTrack~\cite{zhou2020tracking}.
Specifically, we apply two different data augmentations to an image, and use them as the starting and ending frame of a video.
We then interpolate the images and annotations linearly to generate a smooth video for training.

With the synthetic video, we fine-tune the network with the tracking transformer head end-to-end from the single-frame detector.
Our fine-tuning protocol follows DETR~\cite{carion2020end} and uses the AdamW optimizer~\cite{loshchilov2017decoupled}, multiplies the backbone learning rate by a factor of $0.1$, and clamps the gradient norm at $0.1$.
We use a base learning rate of $0.0001$.
We generate video clips of length $T=8$ and train with a batch size of $8$ videos on 8 GPUs, resulting in an effective batch size of $64$.
We fine-tune the network for 22,500 iterations (a $2\times$ schedule).
The fine-tuning takes around 8 hours on 8 Quadro RTX 6000 GPUs.

\par \noindent \textbf{MOT training.}
For our MOT model, we follow past works~\cite{zhou2020tracking,zhang2020fair} to use CenterNet~\cite{zhou2019objects} with a DLA-34 backbone~\cite{yu2018deep} as the object detector.
We use BiFPN~\cite{tan2020efficientdet} as upsampling layers instead of the original deformable-convolution-based~\cite{dai2017deformable} upsampling~\cite{yu2018deep}.
We use RoIAlign~\cite{he2017mask} to extract features for our global tracking transformer.
We do not refine bounding boxes from the RoI feature and use the CenterNet detections as-is.
We use a training size of $1280 \times 1280$ and a test size of $1560$ (longer edge).
Following CenterTrack~\cite{zhou2020tracking}, we pretrain the detector on Crowdhuman~\cite{shao2018crowdhuman} for 96 epochs.
We then fine-tune with the GTR head on Crowdhuman (with augmentation) and the MOT training set in a $1:1$ ratio~\cite{transtrack}.
We again use $T=8$ frames for a video clip with a batch size of $8$ clips.
We fine-tune the network for 32K iterations, which corresponds to $\sim$36 epochs of Crowdhuman and 64 epochs of MOT.
This takes $\sim$6 hours on 8 Quadro RTX 6000 GPUs.

\par \noindent \textbf{Inference.}
During testing, we set the output score threshold to $0.55$ for MOT and the proposal score threshold to $0.4$ for TAO, based on a sweep on the validation set.
We do not set an output threshold for TAO.
For both datasets, we set the new-track association threshold to $\theta=0.2$.
Since the MOT dataset has high frame rate, we find it beneficial to use location information during association.
We associate tracks based on the maximum of the trajectory association score and the box-trajectory IoU.
This is examined in a controlled experiment in \refsec{ablation}.
We further remove trajectories~\cite{braso2020learning} of length~$<5$.

\par \noindent \textbf{Tracking-conditioned classification.}
Our global association module is applied to object features before classification.
This allows us to to classify objects using temporal cues from tracking.
On our TAO experiments, we assign a single classification score to the trajectory by averaging 
the per-box classification scores within the trajectory to a global classification score offline.

\par \noindent \textbf{Runtime.}
We measure the runtime on our machine with an Intel Core i7-8086K CPU and a Titan Xp GPU.
On the MOT17 our backbone detector runs in $47$ms and the tracking transformer in $4$ms per frame. On TAO the backbone runs in $86$ms, the transformer in $3$ms.

\subsection{Global versus local association}
\lblsec{global}
We first validate our main contribution: global association.
We compare to baseline local trackers based on location (SORT~\cite{Bewley2016_sort}), identity, or joint location and identity (FairMOT~\cite{zhang2020fair}).
To make a direct comparison of trackers,
we apply all baseline trackers to the detection outputs of the same model to ensure the same detections (Rows \rowNumber{1-3} in \reftbl{global}).
The ReID features are trained with our association loss (see discussions in \refsec{connection}), we also include baselines with the original instance-classification-based ReID losses (row \rowNumber{4} in \reftbl{global}). 
We adopt the implementation from FairMOT~\cite{zhang2020fair} with the default hyperparameters\footnote{We tuned the hyperparameters, but observed that the default settings performed best.} and tricks,
including a track-rebirth mechanism for up to $30$ frames, for all baselines.

\reftbl{global} shows the results on the TAO~\cite{dave2020tao} and MOT17~\cite{MOT16} validation sets.
First, despite close MOTA and DetA, ReID-based methods (FairMOT~\cite{zhang2020fair} and ours) generally achieve higher tracking accuracy than the location-only baseline~\cite{Bewley2016_sort}.
For our method, when $T=2$ it reduces to a local tracker that only associates across consecutive pairs of frames.
This tracker cannot recover from any occlusion or missing detection, yielding a relatively low AssA.
However, when we gradually increase the temporal window $T$, we observe a consistent increase in association accuracy.
On MOT17 with $T=32$, our method outperforms FairMOT~\cite{zhang2020fair} by a healthy $1.8$ AssA and $1.7$ IDF1, showing the advantage of our global tracking formulation.
On TAO the performance saturates at $T=16$.
This may due to the much lower frame-rate in TAO dataset which results in drastic layout change within a long temporal window.

\begin{table}
\centering
\small
\begin{tabular}{@{}l@{}c@{\ \ }c@{\ \ }c@{\ \ }c@{\ \ }c@{\ \ }c@{\ \ }c@{}}
\toprule
& \multicolumn{4}{c}{Validation} & \multicolumn{1}{c}{Test} & \\ 
  & mAP\@50 & HOTA & DetA & AssA & mAP\@50 & FPS\\ 
\cmidrule(r){1-1}
\cmidrule(r){2-5}
\cmidrule(r){6-6}
\cmidrule(){7-7}
SORT\_TAO~\cite{dave2020tao} & 13.2 & - & - & - & 10.2 & \bf 15.2 \\
QDTrack~\cite{qdtrack} & 16.1 & 35.8 & 24.3 & 53.5 & 12.4 & 5.4 \\
GTR w. QDTrack det. & 20.4 & 40.7 & 30.1 & 55.6 & - & - \\
GTR & {\bf 22.5} & {\bf 45.8} & {\bf 36.8} & {\bf 57.5} & {\bf 20.1} & 11.2 \\
\cmidrule(r){1-1}
\cmidrule(r){2-5}
\cmidrule(r){6-6}
\cmidrule(){7-7}
\textcolor{gray}{AOA~\cite{Du_2020_TAO}} & \textcolor{gray}{25.8} & - & - & - & \textcolor{gray}{27.5} & \textcolor{gray}{1.0} \\
\bottomrule
\end{tabular}
\vspace{-3mm}
\caption{
\textbf{Results on TAO dataset~\cite{dave2020tao}.} We show the HOTA metrics on the validation set and the official metric tracking mAP50. We show the frame-per-second tested on our machine in the last column. 
We show the 2020 TAO challenge winner which are based on a separate ReID network for \emph{per-box} in the last row.
}
\lbltbl{tao}
\vspace{-5mm}
\end{table}

\subsection{Comparison to the state-of-the-art}
\lblsec{sota}
Next we compare to other trackers with different detections on the corresponding test sets.
\reftbl{tao} shows the results on TAO validation and test sets.
TAO~\cite{dave2020tao} is a relatively new benchmark with few public entries~\cite{dave2020tao,qdtrack}.
Our method substantially outperforms the official SORT baseline~\cite{dave2020tao} and the prior best result (QDTrack~\cite{qdtrack}), yielding a relative improvement of 62\% in mAP on the test set.
While part of the gain is from our stronger detector,
this highlights one of the advantages of our model: it is end-to-end jointly trainable with state-of-the-art detection systems.
\reftbl{tao} 3rd row show GTR with the detections from QDTrack~\cite{qdtrack}.
We show GTR displays a $4.3$ mAP and $1.9$ AssA gain over QDTrack using the same detector.

Our model underperforms the 2020 TAO Challenge winner AOA~\cite{Du_2020_TAO},
which trains separate ReID networks on large single-object tracking datasets~\cite{real2017youtube,huang2019got,russakovsky2015imagenet}.
They feed all detected boxes separately to the ReID networks in a slow-RCNN~\cite{girshick2014rich} fashion.
On our machine, AOA's full detection and tracking pipeline takes $989$ms per image on average.
Our model is more than $10\times$ faster than AOA~\cite{Du_2020_TAO}, and uses a single forward pass for each frame with a light-weighted per-object head.

\reftbl{mot17test} compares our tracker with other entries on the MOT17 leaderboard.
Our entry achieves a $74.1$ MOTA, $71.1$ IDF1, and $59.0$ HOTA.
This is better than most concurrent transformer-based trackers, including Trackformer~\cite{meinhardt2021trackformer}, MOTR~\cite{zeng2021motr}, TransCenter~\cite{xu2021transcenter}, and TransTrack~\cite{transtrack}.
Our model currently underperforms TransMOT~\cite{chu2021transmot} in both MOTA and IDF1.
There are several implementation differences between TransMOT and ours, including the use of additional data (TransMOT uses additional ReID data), detector architecture (TransMOT uses YOLOv5~\cite{yolov5} as detector and uses a separate tracker), and training and testing parameters (Code not released).
Our tracker is $1.4$ MOTA lower, but runs $2\times$ faster.

\begin{table*}
\centering
\begin{tabular}{@{}l@{}c@{\ \ \ }c@{\ \ \ }c@{\ \ \ }c@{\ \ \ }c@{\ \ \ }c@{\ \ }c@{\ \ \ }c@{\ \ \ }c@{}}
\toprule
  & MOTA$\uparrow$ & IDF1$\uparrow$ & HOTA$\uparrow$ & DetA$\uparrow$ & AssA$\uparrow$ & FP$\downarrow$ & FN$\downarrow$ & IDS$\downarrow$ & FPS $\uparrow$\\
\midrule
Trackformer~\cite{meinhardt2021trackformer} & 65.0  & 63.9 & - & - & - & 70,443 & 123,552 & 3,528 & -\\
MOTR~\cite{zeng2021motr} & 65.1 & 66.4 & - & - & - & 45,486 & 149,307 & \bf 2,049  & - \\
ChainedTracker~\cite{peng2020chained} & 66.6 & 57.4 & 49.0 & 53.6 & 45.2 & 22,284 & 160,491 & 5,529  & 6.8 \\
CenterTrack~\cite{zhou2020tracking} & 67.8 & 64.7 & 52.2 & 53.8 & 51.0 & \bf 18,498 & 160,332 & 3,039  & 17.5 \\
QDTrack~\cite{qdtrack} & 68.7 & 66.3 & 53.9 & 55.6 & 52.7 & 26,589 & 146,643 & 3,378  & 20.3 \\
TraDeS~\cite{Wu2021TraDeS} & 69.1  & 63.9 & 52.7 & 55.2 & 50.8 & 20,892 & 150,060 &  3,555 & \bf 66.9\\
TransCenter~\cite{xu2021transcenter} & 73.2  & 62.2 & 54.5 & 60.1 & 49.7 & 23,112 & 123,738 & 4,614 & 1.0 \\
GSDT~\cite{wang2020joint} & 73.2  & 66.5 & 55.2 & 60.0 & 51.0 & 26,397 & 120,666 & 3,891 & 4.9\\
FairMOT~\cite{zhang2020fair} & 73.7 & 72.3 & 59.3 & 60.9 & 58.0 & 27,507 & 117,477 & 3,303  & 25.9\\
TransTrack~\cite{transtrack} & 74.5 & 63.9 & 53.9 & 60.5 & 48.3 & 28,323 & 112,137 & 3,663  & 59.2 \\
CSTrack~\cite{liang2020rethinking} & 74.9 & 72.6 & 59.3 & 61.1 & 57.9 & 23,847 & 114,303 & 3,567 & 15.8 \\
FUFET~\cite{shan2020fgagt} & 76.2 & 68.0 & 57.9 & \bf 62.9 & 53.6 & 32,796 & 98,475 & 3,237  & 6.8\\
CorrTracker~\cite{wang2021multiple} & 76.5 & 73.6 & \bf 60.7 & 62.8  & \bf 58.9 & 29,808 & 99,510 & 3,369 & 15.6\\
TransMOT~\cite{chu2021transmot} & \bf 76.7 & \bf 75.1 & - & - & - & 36,231 & \bf 93,150 & 2,346 & 9.6 \\ 
GTR (ours) & 75.3 &  71.5 & 59.1 & 61.6 & 57.0 & 26,793 & 109,854 & 2,859 & 19.6\\ 
\bottomrule
\end{tabular}
\vspace{-3mm}
\caption{
\textbf{Comparison to the state-of-the-art on the MOT17 test set (private detection).} 
We show the official metrics from the leaderboard. $\uparrow$: higher better and $\downarrow$: lower better. FPS is taken from the leaderboard or paper. GTR achieves top-tier performance on MOT17.
}
\vspace{-3mm}
\lbltbl{mot17test}
\end{table*}

\begin{table*}
\centering
\begin{subfigure}{0.49\linewidth}
\centering
\begin{tabular}{@{}l@{}c@{\ }c@{\ \ }c@{\ }c@{}}
\toprule
 & HOTA & DetA & AssA & MOTA \\
\midrule
Direct dot product & 61.3 & 59.5 & 63.6 & 70.5 \\
*Encoder attention  & 63.0 & 60.4 & 66.2 & 71.3 \\
Enc.+ Dec. attention & 62.3 & 60.5 & 64.5 & 71.2\\
\bottomrule
\end{tabular}
\caption{\scriptsize \textbf{With/without attention layers.}
Encoder attention improves tracking.}
\lbltbl{att}
\end{subfigure}
{
\begin{subfigure}{0.50\linewidth}
\centering
\begin{tabular}{@{}l@{}c@{\ }c@{\ \ }c@{\ }c@{}}
\toprule
 & HOTA & DetA & AssA & MOTA \\
\midrule
*no embedding & 63.0 & 60.4 & 66.2 & 71.3 \\
w. positional emb. & 62.5 & 60.7 & 65.0 & 71.7 \\
w.pos.+ temp. emb. & 62.4 & 60.7 & 64.6 & 71.7\\
\bottomrule
\end{tabular}
\caption{\scriptsize \textbf{Different positional/temporal embeddings.} Possitional embeddings do not help.}
\lbltbl{embedding}
\end{subfigure}
}
\begin{subfigure}{0.43 \linewidth}
\centering
\begin{tabular}{@{}l@{\ \ }c@{\ \ }c@{\ \ }c@{\ \ }c@{\ \ }c@{}}
\toprule
 Enc. & Dec. & HOTA & DetA & AssA & MOTA \\
\midrule
*1 & 1  & 63.0 & 60.4 & 66.2 & 71.3 \\
1 & 2 & 62.7 & 60.4 & 65.0 & 71.2 \\
2 & 1 & 63.0 & 60.9 & 66.0 & 71.7\\
\bottomrule
\end{tabular}
\caption{\scriptsize \textbf{Number of transformer layers.} One layer is sufficient for both.}
\lbltbl{layers}
\end{subfigure}
\begin{subfigure}{0.51\linewidth}
\centering
\begin{tabular}{@{}l@{}c@{\ \ }c@{\ \ }c@{\ \ }c@{\ \ }c@{}}
\toprule
& \multicolumn{4}{c}{MOT17} & \multicolumn{1}{c}{TAO} \\ 
 & HOTA & DetA & AssA & MOTA & mAP50\\
\cmidrule(r){1-1}
\cmidrule(r){2-5}
\cmidrule(){6-6}
w/o location & 61.7 & 60.6 & 63.3 & 71.3 & 22.5\\
*w/ location & 63.0 & 60.4 & 66.2 & 71.3 & 22.5\\
\bottomrule
\end{tabular}
\caption{\scriptsize  \textbf{With/without using location during testing.} Location helps MOT17 but not TAO.}
\lbltbl{loc}
\end{subfigure}
\vspace{-2mm}
\caption{\textbf{Design choice experiments on the MOT17 validation set.} * means our default setting. We ablate the effectiveness of attention layers, effectiveness of positional embeddings, number of transformer layers, and use of localizations in testing.}
\vspace{-6mm}
\end{table*}

\subsection{Design choice experiments}
\lblsec{ablation}

Here we ablate our key design choices. All experiments are conducted under the best setting of \reftbl{global}, with $T=32$. The random noise across different runs is within $0.2$ MOTA and $0.5$ AssA.

\par \noindent \textbf{Attention structure.}
We first verify the necessity of using a transformer structure for the association head.
As the counterpart, we remove both the self-attention layer and the cross-attention layer in \reffig{structure}, and directly dot-product the object features after the linear layers. \reftbl{att} shows that this decreases AssA considerably.
Further adding self-attention layers in the decoder as in DETR~\cite{carion2020end} does not improve performance, thus we just use encoder attention.

\par \noindent \textbf{Positional embedding.}
Positional embedding is a common component in transformers. We have implemented a learned positional embedding as well as a learned temporal embedding.
However, we didn't observe an improvement in association accuracy from these, as shown in \reftbl{embedding}.
We thus don't use any positional embedding in our final model.

\par \noindent \textbf{Transformer layers.} 
\reftbl{layers} shows the results of using different numbers of attention layers in the encoder and decoder. While most transformer-based trackers~\cite{transtrack,meinhardt2021trackformer} require 6 encoder and decoder layers, we observe that 1 layer in each is sufficient in our model.
One possible reason is that other trackers take pixel features as input, while we use detected object features, which makes the task easier.

\par \noindent \textbf{Using location in testing.} 
As described in \refsec{inference}, we combine the trajectory probability and location-based IoU during inference.
\reftbl{loc} examines this choice.
On MOT17, using location improves AssA by $3$, due to the high frame-rate on the dataset.
On TAO, where frame-rate is low, using our predicted association alone works fine.

\section{Conclusion}
\lblsec{conclusions}
\vspace{-3mm}
We presented a framework for joint object detection and tracking.
The key component is a global tracking transformer that takes object features from all frames within a temporal window and groups objects into trajectories.
Our model performs competitively on the MOT17 and TAO benchmarks.
We hope that our work will contribute to robust and general object tracking in the wild.

\noindent\textbf{Limitations.} Currently, we use a temporal window size of 32 due to GPU memory limits, and rely on a sliding windows inference to aggregate identities across larger temporal extents.
It can not recover from missing detection or occlusions larger than 32 frames.
In addition, our TAO model is currently trained only on static images, due to a lack of publicly available multi-class multi-object tracking training sets.
Training our model on emerging large-scale datasets such as UVO~\cite{wang2021unidentified} is an exciting next step.

{\small
\par \noindent \textbf{Acknowledgments.}
This material is based upon work supported by the National Science Foundation under Grant No. IIS-1845485 and IIS-2006820.
Xingyi is supported by a Facebook Fellowship.
}

{\small
\bibliographystyle{ieee_fullname}
\bibliography{egbib}
}

\end{document}